%% file: main.tex
\documentclass{article}

\usepackage{hyperref}

\usepackage[accepted]{mlsys2023}
\usepackage{algorithm}
\usepackage{multirow}
\usepackage{comment}
\usepackage{latexsym}
\usepackage{amsmath}
\usepackage{amssymb}
\usepackage{ifthen}
\usepackage{graphicx}
\usepackage{enumitem}
\setlist{nosep}
\usepackage{booktabs}
\usepackage{hyperref}
\usepackage{subcaption}
\usepackage[dvipsnames]{xcolor}
\usepackage[T1]{fontenc}
\usepackage[scaled=0.82]{beramono}
\usepackage{listings}
\usepackage{mathtools}
\usepackage[utf8]{inputenc}
\usepackage{multirow}

\input{std-macros}

\input{macros}

\input{listing-macros}

\DeclarePairedDelimiter\floor{\lfloor}{\rfloor}

\begin{document}

\twocolumn[
\mlsystitle{MGit: A Model Versioning and Management System}
\mlsyssetsymbol{equal}{*}

\begin{mlsysauthorlist}
\mlsysauthor{Wei Hao}{equal,col,msr}
\mlsysauthor{Daniel Mendoza}{equal,stan,msr}
\mlsysauthor{Rafael da Silva}{msr}
\mlsysauthor{Deepak Narayanan}{msr}
\mlsysauthor{Amar Phanishayee}{msr}
\end{mlsysauthorlist}

\mlsysaffiliation{col}{Columbia University}
\mlsysaffiliation{stan}{Stanford University}
\mlsysaffiliation{msr}{Microsoft Research}

\mlsyscorrespondingauthor{Deepak Narayanan and Amar Phanishayee}{dnarayanan@microsoft.com and amar@microsoft.com}

\vskip 0.3in

\input{tex/abstract}
]

\printAffiliationsAndNotice{\mlsysEqualContribution} % otherwise use the standard text.
\input{tex/introduction}
\input{tex/target_settings}
\input{tex/lineage_graph}
\input{tex/graph_construction}
\input{tex/storage}
\input{tex/applications}

\input{tex/evaluation}
\input{tex/related_work}
\input{tex/conclusion}
\bibliography{main}
\bibliographystyle{mlsys2023}

\clearpage
\newpage

\appendix
\input{tex/additional_details}

\end{document}

%% file: std-macros.tex
\ifthenelse{\isundefined{\definition}}{\newtheorem{definition}{Definition}}{}
\ifthenelse{\isundefined{\assumption}}{\newtheorem{assumption}{Assumption}}{}
\ifthenelse{\isundefined{\hypothesis}}{\newtheorem{hypothesis}{Hypothesis}}{}
\ifthenelse{\isundefined{\proposition}}{\newtheorem{proposition}{Proposition}}{}
\ifthenelse{\isundefined{\theorem}}{\newtheorem{theorem}{Theorem}}{}
\ifthenelse{\isundefined{\lemma}}{\newtheorem{lemma}{Lemma}}{}
\ifthenelse{\isundefined{\corollary}}{\newtheorem{corollary}{Corollary}}{}
\ifthenelse{\isundefined{\alg}}{\newtheorem{alg}{Algorithm}}{}
\ifthenelse{\isundefined{\example}}{\newtheorem{example}{Example}}{}

%% file: macros.tex
\newcommand\tldrDone[1]{}

%% file: listing-macros.tex
\lstdefinelanguage{pseudocode}{
 keywords={add_node, add_edge, get_nodes, add_version_edge, sample, federated_averaging, copy_parameters, random_initialization, DataLoader, LineageGraph, cross_entropy_loss_fn, backward, step, traversal, traversal_mtl, run_tests, match, run_update_cascade_vanilla, run_update_cascade_mtl, run_update_cascade, run_function, skip_fn, skip_fn2, terminate_fn, BFS, get_next_version, merged_cr, BFS, DFS, register_test_function, register_creation_function, mgit, diff, add, cr, cr', cr_1, cr_2, cr_n, f_i, f, CreationFunctionFinetuning, CreationFunctionMultiTaskLearning, initialize_model, run_iteration, __call__, next, has_next, isinstance, traversal_all_parents_first, quantize, lcs, compressor, decompressor, dequantize, matched, register_init_from, delta_compression, sort, filter, check, share_parameters, module_diff, generate_hash_table, difference, union, drop, range, min, len, model.save_pretrained, Trainer.save, torch.save, merge, remove_edge, remove_node, deregister_test_function, remove},
 keywordstyle=\color{NavyBlue}\bfseries,
 ndkeywords={for, return, if, else, while, len, def, in, lambda, or, zip, optional, class, list},
 ndkeywordstyle=\color{BrickRed}\bfseries,
 identifierstyle=\color{black},
 sensitive=false,
 comment=[l]{\/\/},
 morecomment=[s]{/*}{*/},
 commentstyle=\color{ForestGreen}\itshape\ttfamily,
 stringstyle=\color{red}\ttfamily,
}

%% file: tex/abstract.tex
\begin{abstract}

\vspace{0.1in}
Models derived from other models are extremely common in machine learning (ML) today. For example, transfer learning is used to create task-specific models from ``pre-trained'' models through finetuning.
This has led to an ecosystem where models are \emph{related} to each other, sharing structure and often even parameter values.
However, it is hard to manage these model derivatives: the storage overhead of storing all derived models quickly becomes onerous, prompting users to get rid of intermediate models that might be useful for further analysis.
Additionally, undesired behaviors in models are hard to track down (e.g., is a bug inherited from an \emph{upstream} model?).
In this paper, we propose a model versioning and management system called MGit that makes it easier to store, test, update, and collaborate on model derivatives.
MGit introduces a lineage graph that records provenance and versioning information between models, optimizations to efficiently store model parameters, as well as abstractions over this lineage graph that facilitate relevant testing, updating and collaboration functionality.
MGit is able to reduce the lineage graph's storage footprint by up to 7$\times$ and automatically update downstream models in response to updates to upstream models.

\end{abstract}

%% file: tex/introduction.tex
\section{Introduction}

ML models are now deployed across a wide set of tasks like vision, language and biology~\citep{he2016deep, devlin2018bert, brown2020language, jumper2021highly}, spanning different target hardware, data and label availability regimes. ML models are served on both specialized accelerators in the datacenter like GPUs and TPUs, as well as on edge devices like mobile phones~\citep{jouppi2017datacenter, murshed2021machine}; similarly, ML models can be trained in a centralized supervised fashion where the training data is available in a single location with associated labels, but also in settings where a large amount of data labels are not available, in settings where the training data is split into disjoint data silos, and in settings where models need to be re-trained daily on new data.

Across these disparate use cases, it has become increasingly common for models to be created that \emph{depend on each other}. For example, when a large amount of supervised data might not be available for a particular task, pretraining can be performed in a self-supervised way on a large \emph{unlabeled} dataset and then these pretrained models can be specialized on a smaller labeled dataset~\citep{pratt1992discriminability, weiss2016survey, bommasani2021opportunities}. For efficient execution on low-powered devices like mobile phones or embedded devices, full-precision models trained on datacenter accelerators are often quantized, pruned and distilled~\citep{polino2018model, guo2021mistify, hao2022tale}. To facilitate model training on independent data silos, new training paradigms like federated learning have been proposed that allow decentralized training~\citep{mcmahan2017communication, bonawitz2019towards}.

\begin{figure*}[t!]
    \centering
    \begin{subfigure}[c]{0.49\textwidth}
        \centering
        \includegraphics[keepaspectratio=1.0,width=0.9\columnwidth]{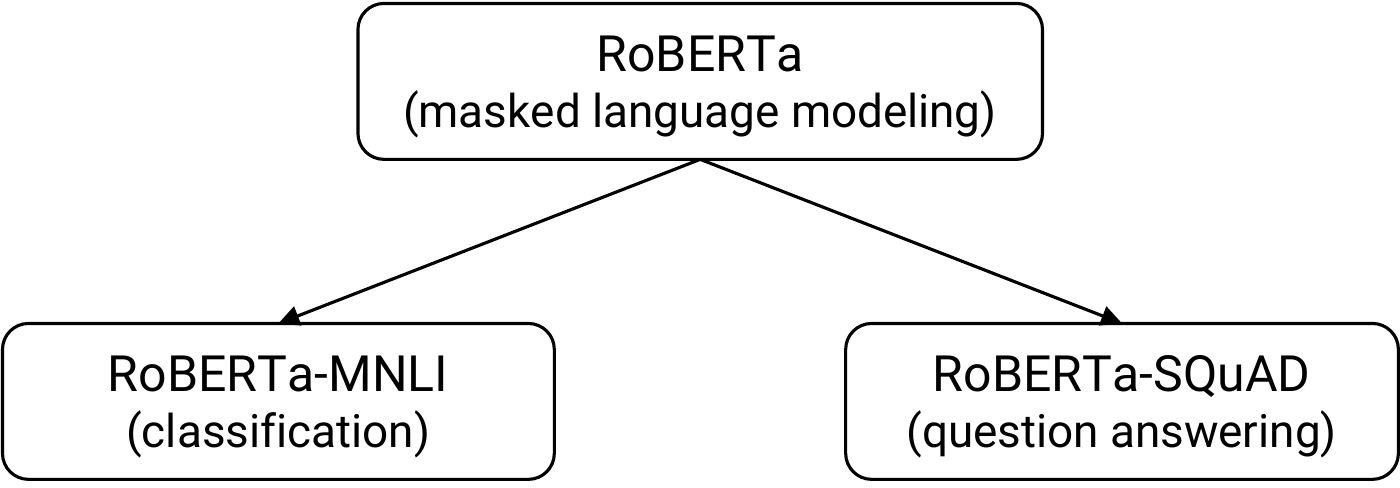}
        \caption{Adaptation.}
        \label{fig:adaptation}
    \end{subfigure}
    \centering
    \begin{subfigure}[c]{0.49\textwidth}
        \centering
        \includegraphics[keepaspectratio=1.0,width=\columnwidth]{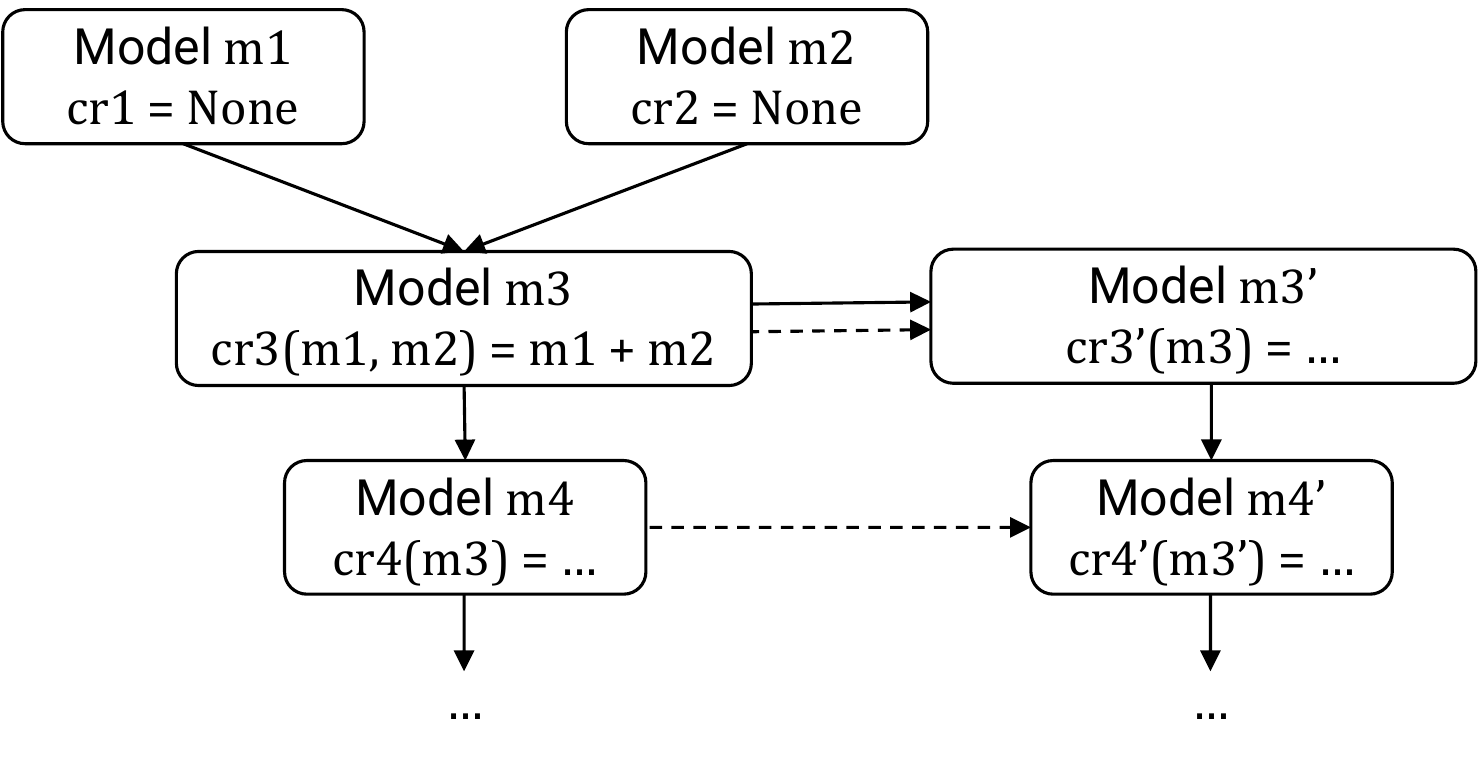}
        \caption{Synthetic example.}
        \label{fig:synthetic}
    \end{subfigure}
    % \begin{subfigure}[c]{0.45\columnwidth}
    %     \centering
    %     \includegraphics[keepaspectratio=1.0,width=\columnwidth]{figures/federated_learning.pdf}
    %     \caption{Federated learning.}
    %     \label{fig:target_settings_federated_learning}
    % \end{subfigure}
    \vspace{-0.05in}
    \caption{
        Example lineage graphs. (a) MGit can be used for various applications with \emph{model derivatives} (e.g., adaptation of models to downstream tasks, federated learning, specialization to edge devices). As an example, we show adaptation. (b) Nodes in the graph are models; a node can be associated with an optional creation function \lstinline$cr$ that specifies how the model can be created from its parent nodes. For example, model \lstinline$m3$ can be created by summing up \lstinline$m1$ and \lstinline$m2$ (contrived example for illustrative purposes). Provenance edges are shown as solid lines, versioning edges are shown as dashed lines. Models are created by following solid edges. Two nodes can have both provenance and versioning edges between them.
    }
    \label{fig:lineage_graph}
\end{figure*}

Despite the widespread use of model derivatives today, no existing system allows for the easy management of these related models. Existing widely-used systems and repositories like PyTorch~\citep{paszke2019pytorch}, TensorFlow~\citep{abadi2016tensorflow} and HuggingFace~\citep{wolf2019huggingface} support the development and management of single models at a time, but dependence information between different models is not stored anywhere.

We believe this is an untapped opportunity. Without automated lineage tracking, various tasks in the model management life cycle are harder and more inefficient:
\begin{itemize}
    \item \textbf{Model storage.} Many models can share parameters exactly, or deviate from ``parent'' models by a small amount, leading to redundancy in storage.
    \item \textbf{Model debugging.} It is hard to debug models that are themselves derived from other models. Does an undesired behavior originate in a given model or in an upstream model?
    \item \textbf{Model updating.} It is hard to update models and keep them in sync. If a certain model is updated, how should dependent models be updated? What if we want to keep certain layers or operators in sync across a collection of models as we update them?
    \item \textbf{Model collaboration.} It is hard for multiple users to \emph{collaboratively} develop models and determine if concurrent changes made to disjoint parts of the model conflict.
\end{itemize}

In this paper, we design and build a system called MGit to make these tasks easier. Our paper makes the following concrete contributions.

\textbf{Lineage graph.} MGit uses a lineage graph data structure to track provenance across multiple ML models through dependency edges, and uses \emph{creation functions} to optionally record how models are created from their parents. MGit's lineage graph also stores other optional metadata like test functions that can be used for model monitoring. A lineage graph can be created automatically from existing model checkpoints, and also manually through a Python or command-line interface.

\textbf{Storage optimizations.} MGit incorporates optimizations to more efficiently store model parameters: it uses content-based hashing and indirection to store parameters shared across models efficiently, and can compress the deltas between non-shared parameters of parent and child models efficiently with no change in underlying model accuracy. MGit's storage optimizations are able to compress model checkpoints by up to 7$\times$ relative to storing each model checkpoint separately.

\textbf{Support for disparate applications.} We also demonstrate a wide set of applications that can use these abstractions to facilitate model testing, diagnostics, updating, and collaborative development. We show that MGit can be used to keep track of dependency information across fine-tuned models, models created using federated learning, and also models specialized for edge devices. Once constructed, the lineage graph can be used to test models and perform diagnostics using a \lstinline$traversal$ primitive. This primitive can also be used to automatically update models given upstream updates. We also provide a \lstinline$merge$ primitive that supports collaboration use cases.

%% file: tex/target_settings.tex
\section{Applications}
\label{sec:applications}

MGit is useful for various applications where models are derived from other models. We describe a few of these in this section; this list is not intended to be exhaustive.

\textbf{Adaptation.}
Transfer learning has been widely adopted to specialize models to various tasks, especially in settings where a large \emph{labeled} dataset might not be available. For example, a model trained with a masked language modeling objective (MLM) using self-supervision could then be finetuned on various text classification tasks with small labeled datasets (Figure~\ref{fig:adaptation}).

\textbf{Model versioning.}
It is often necessary to update models (e.g., to fix an undesired behavior or to re-train models on new training data), creating multiple model versions.

\textbf{Federated learning (FL).}
Federated learning makes it possible to train models in a decentralized way, ensuring that the entire training dataset does not need to be available in a single central location like the cloud. This is advantageous in application settings where privacy concerns might preclude data being uploaded to a central repository or network bandwidth is limited. Instead, models are updated largely locally, and then coalesced periodically to obtain the global shared model.

\textbf{Specialization to edge devices.}
Dense full-precision models often can be too memory- or compute-intensive to run efficiently on edge devices like mobile phones, necessitating techniques like quantization and pruning~\citep{hao2022tale} to compress the model.

\textbf{Multi-task learning (MTL).}
Conventionally, ML models are trained on single tasks. In the multi-task learning (MTL) paradigm, a single model is trained on multiple tasks, with largely shared model parameters and a few task-specific parameters~\citep{ruder2017overview} to improve model generalization.

%% file: tex/lineage_graph.tex
\begin{table*}[t!]
\centering
\begin{tabular}{p{0.14\textwidth}p{0.8\textwidth}}
\toprule
Component & Description \\
\toprule
Node  & ML models are represented in the lineage graph as nodes. Each node has an optional creation function \lstinline$cr$ that tracks how the model can be created from its parents. Nodes are also associated with other relevant metadata like the model type and a unique name. \\
\midrule
Provenance edge & Edge between a model and model(s) derived from it. Tracks how models are created. Can be followed to update models when an upstream model is modified. \\
\midrule
Versioning edge  & Edge between two consecutive versions of the same model. Used to track updates to a model, and can be queried (e.g., to run tests on all model versions). \\
\bottomrule
\end{tabular}
\caption{Components of MGit's lineage graph.}
\label{table:lineage_graph}
\end{table*}

\begin{table*}[h!]
\centering
\begin{tabular}{p{0.35\textwidth}p{0.58\textwidth}}
\toprule
API Name & Description \\
\toprule
\lstinline$add_node(x, xn, [optional] cr)$ & Adds a model \lstinline$x$ as a node to the lineage graph with name \lstinline$xn$. A creation function \lstinline$cr$ can be optionally specified. \\
\lstinline$add_edge(x, y$)  & Adds a provenance edge between nodes \lstinline$x$ and \lstinline$y$. Calls \lstinline$add_node(x)$ and \lstinline$add_node(y)$ if nodes \lstinline$x$ and \lstinline$y$ do not already exist. \\
\lstinline$add_version_edge(x, y)$ & Adds a versioning edge between nodes \lstinline$x$ and \lstinline$y$. \lstinline$x$ and \lstinline$y$ must have the same model type. Calls \lstinline$add_node$ if \lstinline$x$ and \lstinline$y$ do not exist. \\
\lstinline$remove_edge(x, y, type)$ & Removes provenance or versioning (specified by \lstinline$type$) between node \lstinline$x$ and \lstinline$y$.\\
\lstinline$remove_node(x)$ & Removes node \lstinline$x$ and its sub-tree from the lineage graph. Calls \lstinline$remove_edge$ on \lstinline$x$ and all of its parents and all edge types.\\
\midrule
\lstinline$register_creation_function(x, cr)$ & Registers a creation function \lstinline$cr$ for node \lstinline$x$. The creation function specifies how the model \lstinline$x$ should be created from its parents. The creation function can also be used to specify MTL groups. \\
\lstinline$register_test_function(t, tn, [optional] x, [optional] mt)$ & Registers a test \lstinline$t$ with name \lstinline$tn$ either for a specific model \lstinline$x$ or for all models of type \lstinline$mt$ (only one of \lstinline$x$ or \lstinline$mt$ should be specified). \\
\lstinline$deregister_test_function(tn, [optional] x, [optional] mt)$ & De-registers a test with name \lstinline$tn$ either for a specific model \lstinline$x$ or for all models of type \lstinline$mt$ (only one of \lstinline$x$ or \lstinline$mt$ should be specified). \\
\midrule
% \textsc{add\_shared\_parameters}$(m1, m2)$  \todo{Get rid of this?} & ``Tie'' the parameters of modules $m1$ (part of some model $x1$) and $m2$ (part of some model $x2$) together, across model updates. \\
\lstinline$traversal()$  & Returns an iterator of individual nodes or a group of nodes encountered in a traversal. An example traversal is \lstinline$BFS$. \\
\lstinline$get_next_version(x)$ & Returns the next version of model \lstinline$x$ if it exists. \\
\midrule
\lstinline$merge(x1, x2)$ & Try to automatically merge the models pointed to by \lstinline$x1$ and \lstinline$x2$; if not possible, manually request conflict resolution from user. \\
\lstinline$run_tests(i, [optional] re)$ & Runs all registered tests matching the specified optional regex \lstinline$re$ on all nodes returned by the iterator \lstinline$i$. \\
\lstinline$run_function(i, f)$ & Runs function \lstinline$f$ (e.g., compute the parameter norm of the model) on all nodes returned by the iterator \lstinline$i$. \\
\lstinline$run_update_cascade(m, m', skip_fn, terminate_fn)$ & Trigger update cascade as a result of the model update \lstinline$m$ $\rightarrow$ \lstinline$m'$. Nodes are visited once all their parents are visited, starting from \lstinline$m$ with provided skip and termination functions. A new version of a model is created if it has a registered creation function \lstinline$cr$. \\
\bottomrule
\end{tabular}
\caption{MGit API. We show both the lower-level API that can be used to access and mutate the lineage graph directly, as well as higher-level methods that provide more sophisticated functionality.}
\label{table:mgit_api}
\end{table*}

\section{Lineage Graph}

% \begin{figure}[t!]
%     \centering
%     \includegraphics[keepaspectratio=1.0,width=0.5\columnwidth]{figures/lineage_graph.pdf}
%     \caption{
%         An example lineage graph. Each model is a node in the graph, and is associated with an optional creation function \lstinline$cr$ that specifies how the model can be created from its parent nodes. For example, model \lstinline$m3$ can be created by summing up \lstinline$m1$ and \lstinline$m2$. Provenance edges are shown as solid lines, versioning edges are shown as dashed lines. Models are created by following solid edges only. Two nodes can have both provenance and versioning edges between them.
%     }
%     \label{fig:lineage_graph}
% \end{figure}

The main data structure in MGit is the lineage graph (Figure~\ref{fig:lineage_graph}). Table~\ref{table:lineage_graph} shows its components: nodes in the lineage graph are individual models, and edges track provenance and versioning information between models. The lineage graph also tracks other metadata, such as the model type and a unique name, which is useful for testing models, updating them, and also mutating the graph.

\subsection{Interface}

Table~\ref{table:mgit_api} shows both MGit's lower-level API that allows access and mutation of the lineage graph, and the higher-level API that supports more complex functionality on top of the lineage graph such as model testing, per-model function evaluation and automated model updating.

The MGit interface can be used from both command line and Python. The command-line interface is analogous to \lstinline$git$'s command-line interface, and provides an easy way for users to view the lineage graph, run registered tests, etc. The Python interface provides a way to access and mutate the lineage graph, and also perform traversals over the models in it (with functionality that supports running tests and updating models as a part of a traversal). % (e.g., to specify that two models have ``tied'' weights and are part of the same MTL group).
To facilitate both Python and command-line interfaces, changes to metadata are serialized to disk at the end of every operation, and de-serialized at the start of every operation. % Besides supporting lower-level functionality like graph access and mutation, the MGit API also supports higher-level functionality (e.g., model testing layered on top of lineage graph traversals and automated model updating).
We provide more details on the API below.

\subsubsection{Node and Edge Addition}

Lineage graphs have two different types of edges: provenance and versioning edges. Provenance edges track how models are created from each other. A model's creation function has one argument for each of the node's ``provenance parents''. Versioning edges track versions of a given model. For each edge type, every node has a list of adjacent child nodes and a list of adjacent parent nodes. The lineage graph is stored with adjacency lists. MGit's API supports node and edge addition.

Node and edge addition can be directly integrated into larger applications.
For example, a federated learning controller~\citep{konevcny2016federated} averages models trained in a decentralized way across different workers and silos. Individual ``private'' models can be trained on private data independently and then merged to create a global model. By using MGit's Python API, new nodes (and corresponding edges) can be added to the lineage graph as they are created in code (e.g., by finetuning a pretrained model).

\subsubsection{Creation Function}
\label{sec:creation_functions}

Each node is associated with an optional \emph{creation} function \lstinline$cr$ that specifies how the model is created from its direct ancestors. The creation function helps facilitate automated model updating if an upstream model in the lineage is updated.
Creating a new model involves calling the function \lstinline$cr$ with the right arguments, and then calling \lstinline$add_edge$ and / or \lstinline$add_version_edge$ as appropriate.

\textbf{Finetuning and adaptation.}
Finetuning and other light-weight adaptation techniques (e.g., adapters~\citep{rebuffi2017learning}, BitFit~\citep{zaken2021bitfit}) involve initializing a new model from the parent's checkpoint (full checkpoint or partial with parameters for certain layers copied over), and then running training iterations (data wrapped in \lstinline$cr$). We show an example creation function below.

\begin{lstlisting}
class CreationFunctionFinetuning:
  def __init__(self):
    self.lg, self.data_loader = mg.LineageGraph(<filepath>), torch.DataLoader(<filepath>)
    
  def initialize_model(self, parent_list):
    self.child_model, parent_model = Model(), parent_list[0].get_model()
    // Copy parameters from parent_model.
    copy_parameters(parent_model, self.child_model)
    self.child_model.head = initialization(child_head_dimensions) // Initialize head.
    
  def run_iteration(self):
    // Iterate through data using DataLoader.
    batch = next(self.data_loader)
    loss = cross_entropy_loss_fn(self.child_model(batch)) // Forward pass.
    // Backward pass -> Step optimizer.
    loss.backward(); optimizer.step()
    
  def __call__(self, parent_list):
    self.initialize_model(parent_list)
    while self.data_loader.has_next():
      self.run_iteration()
    return self.child_model
\end{lstlisting}

\textbf{Edge device specialization.}
Quantization and pruning also fit into this framework. For example, a simple form of quantization can just downcast each parameter tensor, which is easy to encode in a function \lstinline$cr$. Distillation is similar, but with a more complex creation function \lstinline$cr$.

\textbf{Multi-task learning.} Typically, some parameters of a multi-task model are shared across tasks, while some parameters are local for each task. MGit facilitates multi-task learning by automatically synchronizing updates of shared parameters across models. We can also use creation functions to train models in a MTL fashion using MGit by specifying which parameters are shared in the creation function, as shown below.

\begin{lstlisting}
class CreationFunctionMultiTaskLearning:
  def __init__(self):
    self.lg = mg.LineageGraph(<filepath>)
    self.data_loader = torch.DataLoader(<filepath>)
    
  def initialize_model(self, parent_list):
    self.child_model = Model()
    parent = parent_list[0]
    sibling = parent.get_children()[0]
    if sibling != self:
      // Share parameters with siblings.
      self.lg.share_parameters(sibling.get_model(), self.child_model)
    else:
      // Copy parameters from parent_model.
      self.lg.copy_parameters(parent.get_model(), self.child_model)
    // Randomly initialize head.
    self.child_model.head = random_initialization(child_head_dimensions)
\end{lstlisting}

\subsubsection{Test Functions}
Similarly each node can be associated with test functions. Test functions are registered using the \lstinline$register_test_function$ method, either one model at a time, or for all models of a particular \emph{type}.
% \vspace{-1em}

%  def __call__(self, parent_list):
%    self.initialize_model(parent_list)
%    initialize_all_offspring(parent_list)
%    active_set = get_shared_models(self)
%    while self.data_loader.has_next():
%      for node in active_set:
%        node.run_iteration()
%    return self.child_model

\subsubsection{Traversals}
Traversals are specified as an iterator over the lineage graph nodes. Nodes can be visited in arbitrary orders. Simple example traversals are BFS and DFS. Traversals can also specify the types of edges that should be traversed. For example, to test all versions of a particular model, we could use a traversal that starts at the first version of the given model and then only follows version edges. % On the other hand, to update models, we would use a traversal that only follows provenance edges, since these track how models are created.
More complex traversals like binary search (for test bisections) are also possible using Python generators.

%% file: tex/graph_construction.tex
\subsection{Graph Construction}

We provide two modes for graph construction: manual and automated.

\textbf{Manual construction.}
Manual mode allows users to directly add nodes to the lineage graph and specify provenance for them, using the provided \lstinline$add_node$, \lstinline$add_edge$, \lstinline$add_version_edge$ and \lstinline$register_creation_function$ APIs. These are available both in the command line or in Python (for example, a FL controller implemented in Python can register nodes and edges in code as new models are created, or finetuning code can directly register edges between the parent model and new model).

\textbf{Automated construction.}
\label{sec:automated_construction}
The automated mode enables automatic extraction of dependency information between models. This mode can speed up construction of a lineage graph from a pool of models created outside MGit (e.g., pre-trained models downloaded from the web) by bootstrapping the process without any user annotation.

We use a custom \lstinline[language=pseudocode]$diff$ primitive to compute the differences between two models, both structural (connectivity between layers of the model) and contextual (values of parameters in the model). \lstinline$diff$ makes no assumptions on the model's architecture, and can also be used for dynamic models like MoEs~\citep{fedus2022switch, du2022glam} that use routing layers with learnt parameters, since \lstinline$diff$ only looks at layer parameters and layer connectivity. 
After obtaining both models' DAG representations~\citep{reed2022torch}, \lstinline[language=pseudocode]$diff$ runs a hash-table-based graph matching algorithm, described in the Appendix, that produces the common and different layers and edges between models $A$ and $B$. The output contains the layers and edges to add and remove to produce model $B$ from model $A$.
% This execution graph matching problem is different than the maximum common sub-graph problem as the graphs in our setting are directed and also nodes can have repeated labels.
Structural and contextual diffs are computed by comparing attributes or also parameter \emph{values}.
MGit calculates two scores $d^\text{structural}$ and $d^\text{contextual}$ based on the number of edges in the \lstinline$diff$ output:
\begin{eqnarray}
d^\text{structural} &=& \frac{|\text{edges}^\text{structural}_\text{diff}|}{|\text{edges}^\text{structural}_A| + |\text{edges}^\text{structural}_B|} \nonumber \\
d^\text{contextual} &=& \frac{|\text{edges}^\text{contextual}_\text{diff}|}{|\text{edges}^\text{contextual}_A| + |\text{edges}^\text{contextual}_B|} \nonumber
\end{eqnarray}
For a model \lstinline[language=pseudocode]$x$, MGit locates the model in the graph that has the smallest contextual and then structural divergence score with \lstinline[language=pseudocode]$x$; this node is chosen as the parent of \lstinline[language=pseudocode]$x$. The automated algorithm only adds provenance edges; versioning edges require user annotation. If no model is sufficiently contextually or structurally similar, \lstinline[language=pseudocode]$x$ is added as a root (node with no parents).  We show the runtime scaling of this algorithm in Figure~\ref{fig:average_insertion_time} for large lineage graphs in \S\ref{sec:auto_insertion_time}.

%% file: tex/storage.tex
\section{Storage Optimizations}
\label{sec:storage_optimizations}

A lineage graph with multiple related models can have redundancy in parameter values. Here, we describe two techniques to efficiently store these models, making it more practical to store a large number of dependent models.

\textbf{Content-based hashing.}
We make the observation that many derived models can share parameters. To not redundantly store duplicate copies of these parameters, we use content-based hashing. MGit manages a global hash table that stores the parameters of all models in a lineage graph. The \texttt{SHA-256} hash of each parameter tensor (using both tensor value and its shape) is used as the hash key. 

\begin{algorithm}[t!]
\begin{lstlisting}
def delta_compression(m2, m1, t_thr):
  // m1 and m2 are the parent and child models.
  // We want to compress m1 - m2.
  // t_thr is a user-configurable test accuracy
  // threshold. If the model after compression
  // has an accuracy difference from m1 larger than
  // t_thr, model compression is rejected.

  // First, run LCS to find a mapping between
  // parameters of the same shape.
  (P1, P2) = lcs(m1, m2)
  
  // Calculate quantized deltas between two
  // parameter sets.
  D = quantize(P1, P2)
  
  // Compressor performs lossless compression.
  // Possible options are RLE, LZMA, etc.
  CD, storage_saving = compressor(D)
  if storage_saving < 1:
    return False, None, m2
  else:
    P2' = dequantize(D, P1)
    
    // Restore parameters that are not compressed.
    m2' = m2.difference(P2).union(P2')
    if run_tests([m2']) - run_tests([m2]) < t_thr:
      return False, None, m2
    else:
      return True, CD, m2'
\end{lstlisting}
\caption{Pseudocode for delta compression.}
\label{alg:delta_compression}
\end{algorithm}

\textbf{Delta compression.}
The non-identical parameters of parent and child models might only differ slightly. This motivates the use of compression and decompression of \emph{parameter deltas}, which can be sparse for similar models, for space savings.
Previous work ~\citep{hu2020delta} explored various lossy and lossless compression methods for delta compression and concluded that combining quantization, which converts the delta from a float array to an integer array (lossy), with lossless compression of the subsequent quantized delta works well for many models. MGit extends this approach for delta compression between models in the lineage graph.
% MGit exposes a configurable global compression mode when the user initializes the graph.

One challenge in compressing deltas in MGit is the fact that parent and child models in the lineage graph might not have identical architectures. To circumvent this, we run a longest common subsequence algorithm to compute a mapping between parameters of the model \emph{with the same shape}. For models with the same architecture, this will reduce to parameters of corresponding layers matching with each other. Given a mapping between parameters $p_1$ and $p_2$ of the two models, MGit first computes the delta $\Delta p$ between each pair of parameters and then quantizes $\Delta p$~\citep{hu2020delta}:
\begin{eqnarray}
\Delta p  = p_1 - p_2, \Delta p^{\text{quantized}} = \floor*{\frac{\Delta p}{ 2 \cdot \log(1 + \epsilon)} + 0.5} \nonumber
\end{eqnarray}
MGit then uses \lstinline$compressor$ and \lstinline$decompressor$ modules to losslessly compress $\Delta p^\text{quantized}$. Different lossless compression techniques can be used like RLE~\citep{RLE} and LZMA~\citep{LZMA}; each provides different tradeoffs between compression ratio and runtime (\S\ref{sec:storage_optimization_evaluation}).

$\epsilon$ is a configurable error bound. Larger $\epsilon$ leads to more values in $\Delta p^\text{quantized}$ being driven to 0, contributing to a higher compression ratio after lossless compression, but also reduces the faithfulness of $\Delta p^\text{quantized}$ to $\Delta p$ and introduces larger accuracy drops. We use a default $\epsilon=10^{-4}$.

MGit only \emph{accepts} delta compression if the compression results in storage saving and an accuracy drop within a configurable threshold (if tests are registered). Each delta-compressed parameter will be stored on disk as the compressed delta along with a pointer to the parent layer to facilitate future decompression. If not, compression is rejected and the uncompressed model is persisted.
This procedure can be applied recursively. That is, the delta can be computed between the layers of a child model and a parent model that is itself delta compressed. Loading a model instance then involves recursively decompressing up the chain until the first ancestor node that is not delta compressed. Full pseudocode is shown in Algorithm~\ref{alg:delta_compression}.

%% file: tex/applications.tex
\section{Higher-Level Functions over Lineage Graph}

\begin{algorithm}[t!]
\begin{lstlisting}
def run_update_cascade(m, m', skip_fn, terminate_fn):
  // First, create (empty) next versions of models.
  skip_fn2 = lambda x: skip_fn(x) or x == m
  for x in BFS(m, skip_fn2, terminate_fn):
    // Get next version of each parent of x if it exists, otherwise get current version.
    ps' = [get_next_version(p) for p in x.parents]
    x' = x.cr.initialize(ps')
    
    // Add provenance and version edges, and copy creation function.
    add_edge(p', x') for p' in ps'
    add_version_edge(x, x')
    x'.cr = x.cr
  
  // Next, start traversal at children of m', and train models by calling creation
  // function. traversal_all_parents_first returns an iterator over nodes (or group
  // of nodes if using MTL) such that a node is visited only once _all_ of its
  // parents (parent MTL groups if using MTL) are visited.
  skip_fn2 = lambda x: skip_fn(x) or x == m'
  for xs' in traversal_all_parents_first(m', skip_fn2, term_fn):
    if isinstance(xs', list):
      // Run MTL using combined creation function.
      merged_cr(xs', xs'.parents)
    else:
      // Otherwise, call individual model node's creation function.
      [x'] = xs'
      x'.cr(x'.parents)
    
\end{lstlisting}
\caption{Pseudocode for model updating.}
\label{alg:model_updating}
\end{algorithm}

We now describe higher-level functions that can be performed over the lineage graph.

\textbf{Testing.}
Given an already constructed lineage graph, MGit exposes APIs to examine models in the graph.
Testing can be thought of as executing per-node test functions \lstinline$t$ as part of a graph traversal. MGit provides a way to register functions both for individual models and for all models of a specific type. Users can specify a regex \lstinline$re$; for every node encountered in the traversal, all registered tests whose names match \lstinline$re$ are run. Running the same test for multiple related models allows users to track model regressions more easily (e.g., all descendents of a particular model might show poor performance on a particular test) and correlate dependency information with model performance on various tasks.
% \begin{lstlisting}
% def run_tests(i, optional re):
%   // Iterator i is the output of a traversal function. Two possible options for traversal
%   // are BFS and DFS, but more complicated traversals are also possible. re is an
%   // optional test regex.
%   for x in i:
%     for t in match(x.tests, re):
%       t(x) // Call t on each visited node x.
% \end{lstlisting}
We can also execute other per-model functions as diagnostics. For example, we could compute the deltas between every model and its parent(s), or measure models' sparsity levels.

\textbf{Model updating.}
Users can notify MGit that a model has been updated (with the checkpoint of the new model). This will then automatically trigger the update cascade on downstream dependent models.
In the most basic form of model updating, when a new version of a model is created, provenance edges in the lineage graph are followed to produce a new set of model versions for all of the model's descendants. We use a modified BFS traversal, where a node is visited only once all of its parents have been visited, to ensure that the creation function is called only when all required upstream models are available. A \emph{new version} of the model is computed using the node's registered \lstinline$cr$ function; MGit never overwrites existing models with its automated functionality since users might want to vet new models. MGit's storage optimizations ensure that multiple versions of the same model can be served with minimal overhead. Full pseudocode is in Algorithm~\autoref{alg:model_updating}.

We can use MTL to continuously share parameters across models even across updates if so desired by using an appropriate creation function (\S\ref{sec:creation_functions}). The traversal to re-train models needs to ensure that full MTL groups are executed only once all MTL groups on which they depend also complete. Additionally, individual creation functions \lstinline$cr$ are not called. Instead, all desired functions \lstinline$cr_1, cr_2, ..., cr_n$ are merged into \lstinline$cr'$ that returns $n$ new models. Internally, this merged creation function \lstinline$cr'$ ensures that weights are shared, appropriate loss functions are used, etc.

\textbf{Collaboration.}
MGit also supports collaboration workflows through a \lstinline$merge$ primitive. The objective of this primitive is to identify if concurrent changes made to a given model are ``compatible'' or not. Changes made to the same layers of a given model need to be manually merged (akin to a manual merge in the case of a merge conflict in \lstinline$git$). Changes made to different layers of a model are also not necessarily compatible: in cases of a dependency between two layers (i.e., one layer consumes the output of the other eventually or a downstream layer consumes the outputs of both layers) changed by different users, additional tests are needed to verify that the concurrent changes did not result in a model regression.

\begin{figure}[t!]
    \centering
    \includegraphics[keepaspectratio=1.0,width=0.95 \columnwidth]{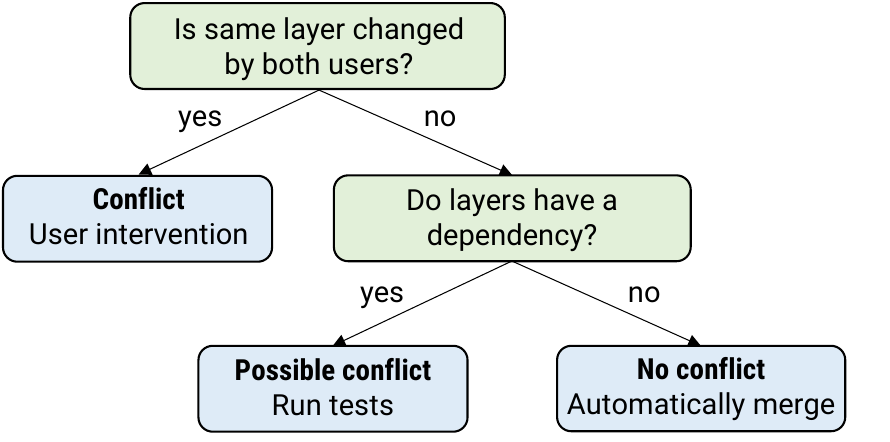}
    \caption{
        Decision tree for merging changes. If a layer is changed by both users, manual merging is required because of the conflict. Otherwise, if changed layers have a dependency, a conflict is possible and tests are required to verify whether this is the case. If neither of the above conditions is true, the merge can be done automatically.
    }
    \label{fig:merge_decision_tree}
\end{figure}

MGit's \lstinline$merge$ primitive helps support collaboration use cases, where multiple users might make ``edits'' to the same model concurrently. \lstinline$merge$ is given two models (models created by concurrent edits) and their closest common ancestor (the original model on which changes were made concurrently) in the lineage graph as input. It returns three possible results:
\begin{itemize}
    \item \textbf{Conflict.} At least one common layer is updated by both changes. In this case, manual intervention is required.
    \item \textbf{Possible conflict.} Two layers changed by different users have a ``dependency''. Consequently, additional tests are needed to verify that the concurrent changes did not result in a model regression.
    \item \textbf{No conflict.} No common layer updated by both users, and no dependency between the any two of the users' changes. In this case, the merge can be processed automatically.
\end{itemize}

The decision tree shown in Figure \ref{fig:merge_decision_tree} summarizes the conflict detection approach implemented in the \lstinline$merge$ primitive. Let \lstinline$m1$ and \lstinline$m2$ be two changed models performed by different users on model \lstinline$m$. Then the above checks can be performed by first computing \lstinline$d1 = diff(m, m1)$ and \lstinline$d2 = diff(m, m2)$, and then performing a DFS through the models to check for dependencies between the changed layers. If such a dependency exists, then the changes are flagged as ``possible conflict'', otherwise the changes commit.

%% file: tex/evaluation.tex
\section{Evaluation}

In this section, we evaluate MGit's storage optimizations, and its ability to enable functionality that would be hard to execute without recording lineage between models. Unless otherwise noted, experiments were run on a workstation with 4 NVIDIA RTX A6000 GPUs and CUDA 11.7.

\subsection{ Lineage Graphs}

\begin{table*}[t!]
\centering
\begin{tabular}{lp{0.23\textwidth}p{0.47\textwidth}c}
\toprule
Name & Graph type & Description & \# Nodes / \# Edges \\
\toprule
$G1$ & HuggingFace & NLP models downloaded from HuggingFace. &  23 / 21 \\
\midrule
$G2$ & Adaptation & BERT-style models specialized for NLP tasks using finetuning and other lightweight adaptation techniques. Some models are trained using different datasets, creating multiple versions. &  91 / 171\\
\midrule
$G3$ & Federated learning & Vision models trained in a decentralized fashion using FL. &  60 / 95\\
\midrule
$G4$ & Edge device specialization & Vision models with pruned model weights for edge devices. &  22 / 19\\
\midrule
$G5$ & Multi-task learning & BERT-style models specialized for NLP tasks with MTL to enforce parameter sharing. & 10 / 9\\
\bottomrule
\end{tabular}
\caption{Lineage graphs considered in evaluation.}
\label{table:lineage_graphs}
\end{table*}

Table~\ref{table:lineage_graphs} shows the lineage graphs considered in this evaluation, reflecting various applications that create ML model derivatives (\S\ref{sec:applications}). $G1$ was automatically constructed using the algorithm outlined in \S\ref{sec:automated_construction}. Graphs $G2$ through $G5$ were manually created using the \lstinline$add$ functions, in conjunction with the training APIs used to create the models.

\textbf{G1.} $G1$ is a lineage graph created from NLP models downloaded directly from the HuggingFace model hub~\citep{huggingface_model_hub}. The full list of models used is:
\begin{itemize}
  \item \texttt{bert-base-cased}.
  \item \texttt{bert-base-uncased}.
  \item \texttt{aloxatel/bert-base-mnli}.
  \item \texttt{ericRosello/bert-base-uncased-finetuned-} \texttt{squad-frozen-v2}.
  \item \texttt{deepset/bert-base-uncased-squad2}.
  \item \texttt{bert-large-uncased}.
  \item \texttt{bert-large-cased}.
  \item \texttt{TehranNLP-org/bert-large-mnli}.
  \item \texttt{roberta-base}.
  \item \texttt{deepset/roberta-base-squad2}.
  \item \texttt{textattack/roberta-base-MNLI}.
  \item \texttt{roberta-large}.
  \item \texttt{roberta-large-mnli}.
  \item \texttt{deepset/roberta-large-squad2}.
  \item \texttt{albert-base-v2}.
  \item \texttt{twmkn9/albert-base-v2-squad2}.
  \item \texttt{prajjwal1/albert-base-v2-mnli}.
  \item \texttt{distilbert-base-uncased}.
  \item \texttt{distilbert-base-cased}.
  \item \texttt{twmkn9/distilbert-base-uncased-squad2}.
  \item \texttt{ericRosello/distilbert-base-uncased-} \texttt{finetuned-squad-frozen-v2}.
  \item \texttt{google/electra-small-generator}.
  \item \texttt{howey/electra-small-mnli}.
\label{table:G1_model_list}
\end{itemize}
We then ran MGit's automated graph construction method on these models to create a lineage graph. 22 out of 23 nodes are correctly inserted relative to a ``gold'' lineage graph. The only mis-inserted model is \texttt{bert-base-uncased}. The automated graph construction function is able to correctly insert models, including some that have frozen weights inherited from their parent model, by computing divergence scores between model pairs. MGit's API allows errors made by the automated algorithm to be corrected manually by users (using the \lstinline$remove$ functions in the API).

\textbf{G2.} We started with a vanilla RoBERTa model trained on the standard masked language modeling (MLM) objective, and then finetuned task-specific models for each of the GLUE tasks~\citep{wang2018glue}. We created 10 versions of each task-specific model by finetuning on additional perturbed data~\citep{moradi-samwald-2021-evaluating}.

\textbf{G3.} We trained a ResNet-50 image classification model~\citep{he2016deep} on the ImageNet-1K dataset~\citep{deng2009imagenet} using federated learning. Each worker operates on a data silo with a subset of the 1000 labels in the ImageNet-1K dataset. We ran experiments with 40 workers (data silos), and 10 rounds of federated averaging. In each round, 5 of 40 workers are randomly sampled.

\textbf{G4.} To create models that can be deployed on the edge, we pruned three image classification models to varying degrees: ResNet-50, DenseNet121~\citep{huang2017densely} and MobileNet-v3~\citep{howard2017mobilenets}. For each model architecture, we create models progressively greater sparsities in a two-step process.
In the first step, a model with sparsity $s_i$ is created by masking out the $s_i$ fraction of its non-zero parameters with lowest magnitude. We then check if the resulting model is accurate enough, and if not, we finetune the model on ImageNet-1K to further improve accuracy while preserving its sparsity.

\textbf{G5.} We use MTL to create RoBERTa models for GLUE tasks with shared weights. This is similar to $G2$.

% We believe this dataset of lineage graphs will be a valuable resource to the community and we will open source it.

\subsection{Auto-Insertion}
\label{sec:auto_insertion_time}

\begin{figure}[t!]
    \centering
    \includegraphics[keepaspectratio=1.0,width=0.8\columnwidth]{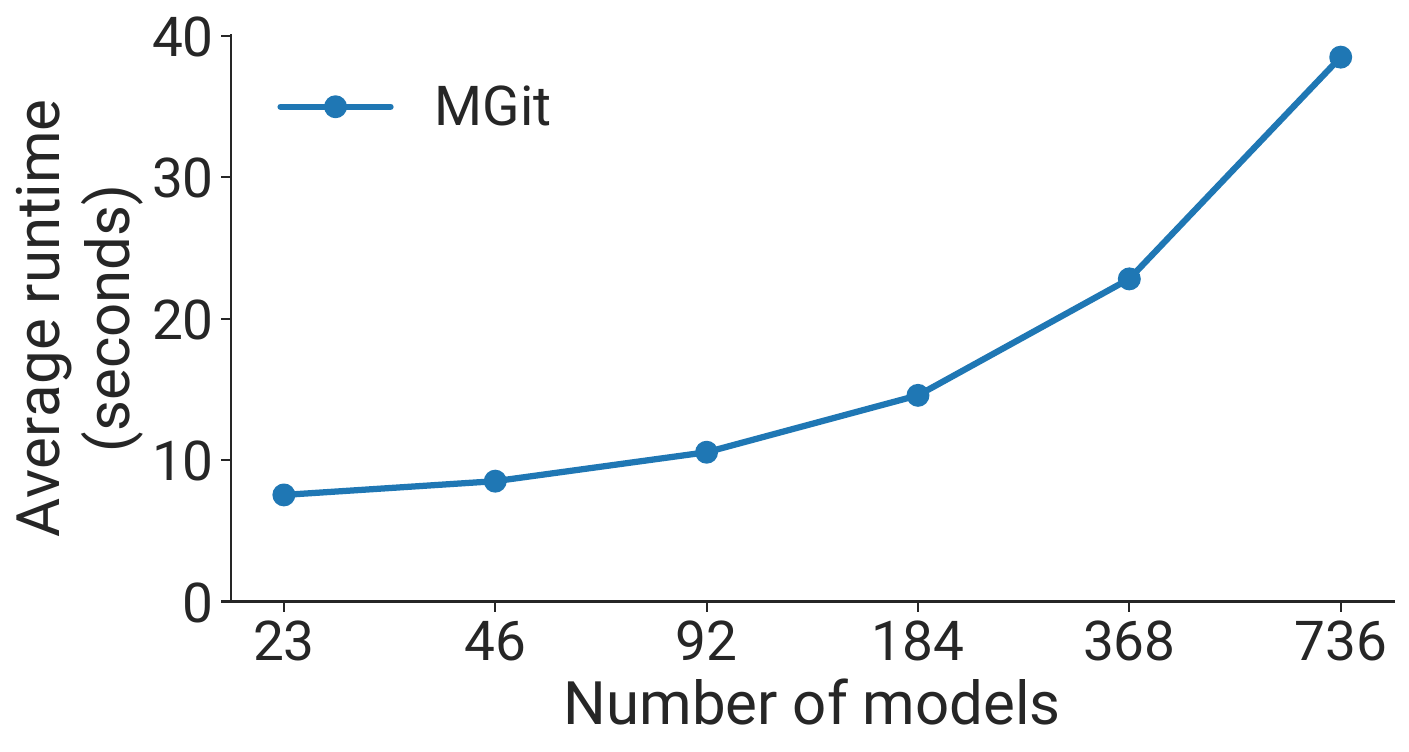}
    \caption{
        Average per-model insertion time for lineage graphs of different sizes.
    }
    \label{fig:average_insertion_time}
\end{figure}

Figure~\ref{fig:average_insertion_time} shows the average per-model insertion time for lineage graphs of different sizes when using the auto-insertion algorithm described in \S\ref{sec:automated_construction}. We create larger graphs by scaling up $G2$ by a desired factor: for example, our graph with 92 models or nodes is created by replicating each model in $G2$'s model pool 4 times. ``Auto-inserting'' a model into the lineage graph involves a pairwise comparison with all other models already in the lineage graph; consequently, the average per-model runtime increases as the lineage graph becomes larger. We believe that with large lineage graphs with hundreds of models, average insertion times of 40 seconds / model are reasonable, especially when compared to the significant times required to train models end-to-end.

\begin{table*}[t!]
\centering
\begin{tabular}{llcccc}
\toprule
Graph & Compression technique & Comp. ratio ($\uparrow$) & \multicolumn{2}{c}{Accuracy $\Delta$ ($\downarrow$)} & Per-model runtime ($\downarrow$) \\
      &                       &             & Max. & Avg. & \\
\toprule
\multirow{5}{*}{$G1$} & \texttt{MGit (LZMA + Hash)} & \textbf{2.14} & 0.09  & 0.01 &  35.7 mins \\
 & \texttt{MGit (RLE + Hash)} & 1.13  & 1.02 & 0.08 & 30.9 mins  \\
 & \texttt{MGit (Hash)} & 1.05 &  \textbf{0.00}& \textbf{0.00} &  \textbf{12.0 mins} \\
 & \texttt{Full} & 1.83 & 0.08& 0.00 &  36.5 mins \\
 & \texttt{Full w/o quantization} & 0.87 & \textbf{0.00}& \textbf{0.00} &  29.8 mins \\
\midrule
\multirow{5}{*}{$G2$} & \texttt{MGit (LZMA + Hash)} & \textbf{5.35} & 0.01& 0.00& 7.4 mins \\
 & \texttt{MGit (RLE + Hash)}  & 1.84 & 0.01 & 0.00 & 4.1 mins \\
 & \texttt{MGit (Hash)} & 1.01 & \textbf{0.00}&\textbf{0.00} & \textbf{0.1 min} \\
 & \texttt{Full} & 1.85 & 0.00 & 0.00 & 14.6 mins \\
 & \texttt{Full w/o quantization} & 0.78 & \textbf{0.00}& \textbf{0.00} & 3.8 mins \\
\midrule
\multirow{5}{*}{$G3$} & \texttt{MGit (LZMA + Hash)} & \textbf{6.96} & 0.11& 0.01& 2.5 mins \\
 & \texttt{MGit (RLE + Hash)} & 3.11 & 0.49& 0.03& 2.4 mins \\
 & \texttt{MGit (Hash)} & 1.00 & \textbf{0.00}& \textbf{0.00}& \textbf{1.1 mins} \\
% $G3$ & (sparse) & 1.02 & 0.17 & 1.2 mins \\
 & \texttt{Full} & 2.29 & 0.25&0.06 & 4.0 mins \\
 & \texttt{Full w/o quantization} & 0.72 & 0.06& 0.01& 2.8 mins \\
\midrule
\multirow{5}{*}{$G4$} & \texttt{MGit (LZMA + Hash)} & \textbf{2.57}  & 0.35& 0.07 & 2.5 mins \\
 & \texttt{MGit (RLE + Hash)} & 2.04  & 0.35& 0.07& 2.5 mins \\
 & \texttt{MGit (Hash)} & 1.00 & \textbf{0.00}& \textbf{0.00} & \textbf{1.1 mins} \\
% $G4$ & (dict) & 1.55  & 0.35 & 3.3 mins \\
 & \texttt{Full} & 2.57 & 0.37& 0.07& 3.0 mins \\
 & \texttt{Full w/o quantization} & 1.47 & 0.07& 0.01& 2.6 mins \\
\midrule
$G5$ & \texttt{MGit (Hash)} & 4.93 & \textbf{0.00}& \textbf{0.00} & \textbf{0.1 min} \\
% $G5$ & \texttt{Full} & \textbf{9.2} & &  & 10.2 mins \\
%  & \texttt{Full w/o quantization} & 4.7 & &  & 4.0 mins \\
\bottomrule
\end{tabular}
\vspace{0.05in}
\caption{Compression ratio, maximum / average accuracy delta across models in lineage graph, and per-model runtime of delta compression techniques for various lineage graphs. \texttt{Full} is the approach of using quantization and LZMA on full models instead of the deltas.}
\label{table:storage_optimization_evaluation}
\end{table*}

\subsection{Storage Optimization}
\label{sec:storage_optimization_evaluation}
We now evaluate MGit's storage optimizations. Table~\ref{table:storage_optimization_evaluation} shows the results for various MGit configurations, combining the content-based hashing and delta compression techniques described in \S\ref{sec:storage_optimizations}. We show results for two versions of the delta compression algorithm: one that uses \texttt{LZMA} for its lossless \lstinline$compressor$ / \lstinline$decompressor$, and another that uses \texttt{RLE} instead. We also show the content-based hashing technique alone (\texttt{Hash}). For $G4$, we quantize parameters before calculating deltas so that the sparsity is preserved in each model. Additionally, we implemented two baselines that run LZMA on either a quantized version or the original full model (\texttt{Full} and \texttt{Full w/o quantization}). We show three metrics: compression ratio (larger is better), maximum accuracy delta between original uncompressed models and models in the compressed lineage graph (smaller is better), and average compression + testing runtime per model (smaller is better).

As a lossless storage method, content-based hashing shows storage savings proportional to the number of parameters duplicated across models in the lineage graph. We observe that these numbers are $9.4\%$, $16.5\%$ and $79.6\%$ for $G1$, $G2$ and $G5$ respectively. $G5$ has more duplicated parameters since its models were explicitly trained to share parameter values using MTL.

For delta compression methods which are lossy due to the quantization step, LZMA shows the best compression ratio across all graphs. % LZMA shows a larger accuracy drop for $G3$ compared to RLE because it compressed more parameters (leading to higher quantization error): in $G3$, the fraction of parameters compressed by LZMA and RLE are $91.7\%$ and $83.3\%$ respectively.

Quantization and LZMA applied to the full models has worse compression ratios than the default MGit approach (compressing deltas) except for $G4$. There are three reasons for this: first, the fraction of compressed parameters in $G4$ is lower compared with $G2$ and $G3$ due to accuracy check failures. Second, the magnitude of deltas in $G4$ is larger than the deltas in other graphs on account of how the models were derived from each other (L1 pruning instead of finetuning). Third, the three roots models in $G4$ were not compressed in MGit whereas they were in the \texttt{Full} baseline. This is an optimization that can be added to MGit.

Lastly, we notice that methods with larger compression ratios often have longer runtimes, but we believe an average runtime of even 10-15 minutes per model is reasonable given long model training times. $G1$ takes particularly long because we ran tests on CPUs as opposed to on GPUs.

\begin{figure}[t!]
    \centering
    \begin{subfigure}[c]{0.95\columnwidth}
        \centering
        \includegraphics[keepaspectratio=1.0,width=0.9\columnwidth]{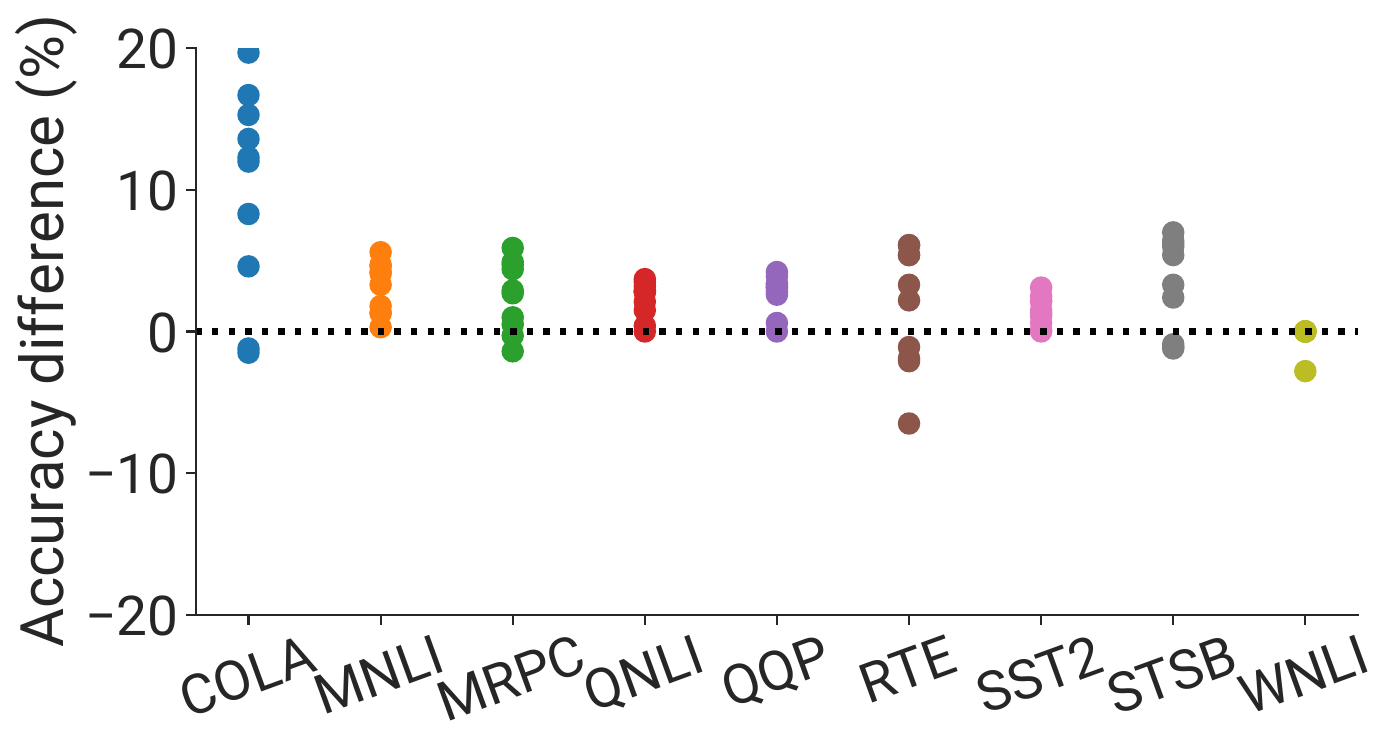}
        \caption{$G2$.}
    \end{subfigure}
    \centering
    \begin{subfigure}[c]{0.95\columnwidth}
        \centering
        \includegraphics[keepaspectratio=1.0,width=0.9\columnwidth]{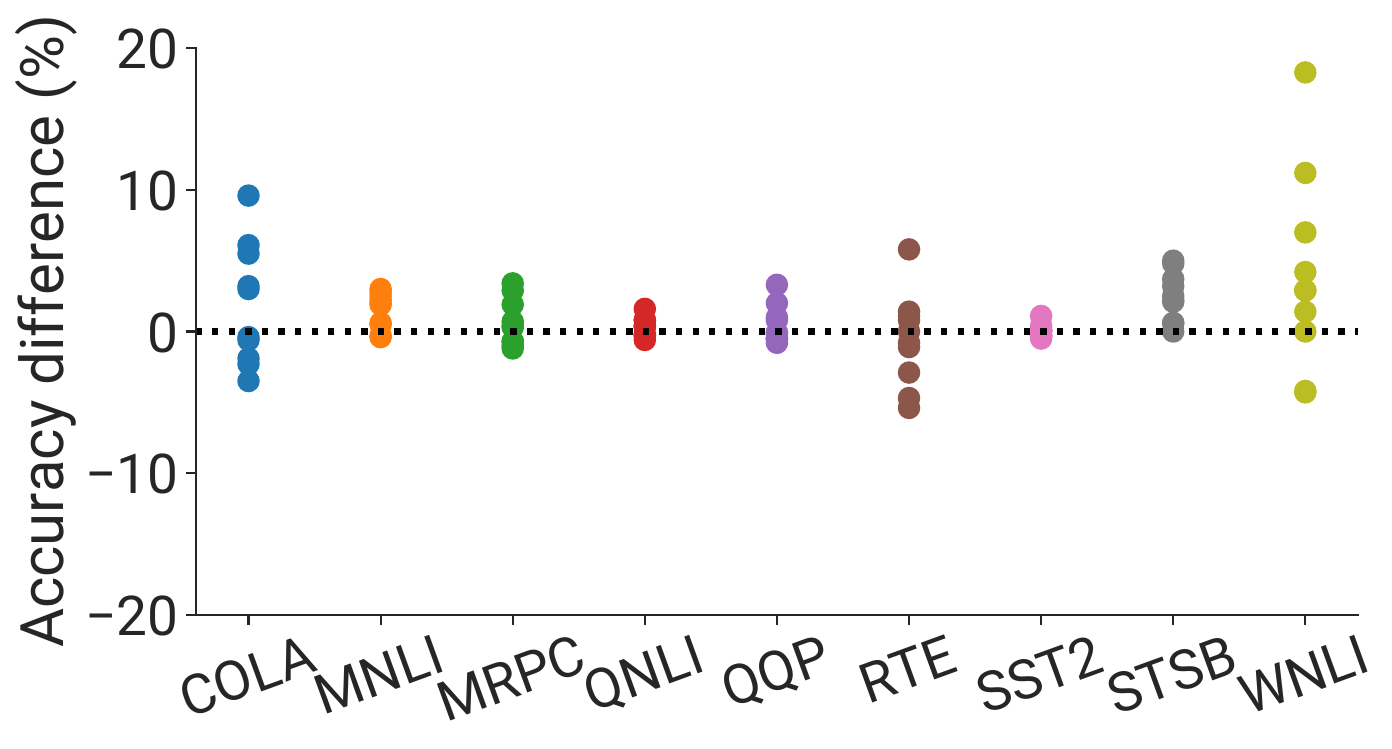}
        \caption{$G5$.}
    \end{subfigure}
    \caption{
        Accuracy difference between models produced by MGit's automated model updating feature and base models for various GLUE tasks.
    }
    \vspace{-0.1in}
    \label{fig:model_updating}
    \vspace{-0.1in}
\end{figure}

\subsection{Functionality}
\label{sec:functionality}

MGit enables functionality that is hard to perform without a model management system.

We found MGit to be useful in testing models by providing a way to combine dependency information with testing functions. For example, MGit facilitates running test bisections, searching for the first model in a lineage chain which fails a particular test. In the best case, we found that failing models can be found as much as 1.5$\times$ faster using test bisections. We expect this runtime improvement to be larger for deeper lineage chains where asymptotic improvements matter more.

MGit is also able to leverage its lineage graph and creation functions to train models that share state. $G5$ was trained using MTL (\S\ref{sec:creation_functions}) to create RoBERTa models for 9 different GLUE tasks~\citep{wang2018glue}; the models in $G5$ shared 98\% of their parameters (only parameters in the model heads were not shared).

We also evaluate MGit's automated model updating functionality (\lstinline$run_update_cascade$ API).
For $G2$ and $G5$, we try to more efficiently build models for each task resilient to various perturbations by finetuning the parent MLM model (\lstinline$m$) with perturbed data, thus generating a new model \lstinline$m'$. We then run \lstinline$run_update_cascade$ to generate new children \lstinline$m1'$, \lstinline$m2'$, $\ldots$, \lstinline$m10'$ from \lstinline$m'$ while reusing the creation functions that facilitated the creation of \lstinline$m1$, etc. from \lstinline$m$. These creation functions do not use perturbed data at all; any ability of \lstinline$m1'$, \lstinline$m2'$, $\ldots$, \lstinline$m10'$ to perform well on the perturbed GLUE tasks will be passed down from \lstinline$m'$.
Figure~\ref{fig:model_updating} shows the accuracy differences between the new models \lstinline$m1'$, \lstinline$m2'$, $\ldots$, \lstinline$m10'$ and the original models \lstinline$m1$, \lstinline$m2$, $\ldots$, \lstinline$m10$  for all GLUE tasks and data perturbations. For most perturbations and GLUE tasks, MGit shows superior performance (accuracy difference $> 0$).

%% file: tex/related_work.tex
\section{Related Work}

We now briefly discuss other work related to MGit and the model management problem it tackles.

\textbf{Adaptation.}
Transfer learning has been used to wide effect to adapt models to various tasks, especially in regimes where a large amount of supervised data might not be available. The most widely adopted model adaptation approach is full finetuning, where all parameters of a pre-trained model are updated while running forward and backward passes on task-specific inputs. To reduce storage requirements when adapting a model to a new task, many lightweight adaptation techniques have been introduced. Instead, of updating the entire model, \citet{Houlsby2019ParameterEfficientTL} introduces adapter modules (small MLPs) within the model architecture (freezing all other weights); this approach ensures that the only new task-specific parameters are in the adapter modules, which are small. BitFit~\citep{ben-zaken-etal-2022-bitfit} proposes only updating the bias vectors for adaptation to a new task. Diff pruning~\citep{guo-etal-2021-parameter} adds sparse vectors to the parameters, while LoRA ~\citep{hu2022lora} adds low-rank vectors to the model parameters. Given this rapid proliferation of lightweight adaptation techniques, PetS~\citep{pets} proposes a general abstraction for lightweight adaptation with the goal of improving \emph{serving efficiency} of transformer models creating using such techniques. MGit provides a convenient abstraction to manage, develop and store lineage of lightweight-adapted models; leveraging lineage for inference runtime improvements is interesting future work.

\textbf{Debugging in ML.}
Testing and debugging is of significant importance when deploying machine learning models. Checklist ~\citep{checklist} observed many state-of-the-art models often exhibit bugs and make incorrect predictions; to help reduce these, Checklist proposes templates and abstractions for manually developing tests for NLP models. AdaTest ~\citep{adatest} proposes a human-in-the-loop test generation solution that uses a ML model to recommend test cases for NLP models. MLEXray ~\citep{mlexray} observes that models deployed at the edge can perform poorly unexpectedly and proposes a system to facilitate monitoring and debugging models at the edge. Model assertions~\citep{kang2020model} identifies various invariants that should hold for model outputs corresponding to related inputs (e.g., frames of a video stream) and proposes ways to correct for any discrepancies that might occur. MGit combines such per-model testing solutions with the notion of a lineage graph, helping developers study failures \emph{across} different models.

\textbf{Model patching.}
After undesired behaviors are identified, models need to be subsequently updated. Cheap model patching approaches have become a target of much study, since training a model from scratch is often prohibitively expensive and time-consuming. This is a challenging problem since we often want to keep such changes \emph{scoped}, making sure to only update the behavior intended while not changing already correct predictions. AdaTest ~\citep{adatest} provides a mechanism to generate new training data with a human-in-the-loop for bug fixing in NLP models. External learned editors~\citep{model_editor,mitchell2021fast,hase} can be used to modify raw finetuning gradients to scope changes. MGit provides a framework to automatically update models that might depend on a buggy model, obviating the need to manually update them.

\textbf{Model repositories.}
ModelHub~\citep{miao2016modelhub} also suggests the usefulness of interacting with ML models using a \lstinline$git$-like interface. However, ModelHub is not intended for derived ML models, and also does not present solutions for automated model updating, testing, and collaboration.
HuggingFace Model Hub~\citep{huggingface_model_hub, wolf2019huggingface} is a widely-used model repository where users can upload their trained models; however, it does not record provenance information.

% Constrained finetuning: Sotoudeh \& Thakur, 2019; Zhu et al., 2020.
% Methods using special pre-training objectives via meta learning: Sinitsin et al., 2020.

%% file: tex/conclusion.tex
\section{Conclusion}

Models are increasingly derived from other machine learning models as the number of deployment settings of ML rises. This greatly complicates modern ML workflows: diagnosing and updating models is more challenging than ever on account of these dependencies. In this paper, we propose a system called MGit that tries to easen this burden using a lineage graph that records dependencies between models and abstractions over the lineage graph that facilitate easier testing, updating and collaboration. MGit's storage optimizations reduce the model storage footprint by up to 7$\times$.

%% file: tex/additional_details.tex
\twocolumn{

\section{Pseudocode for \lstinline$diff$ Primitive}
\label{sec:appendix_additional_pseudocode}

\lstinline$diff$ is a key MGit primitive that powers its automated graph construction algorithm. Algorithm~\ref{alg:graph_diff} shows its implementation.

\input{tex/diff_pseudocode}

}

%% file: tex/diff_pseudocode.tex
\label{sec:graph_diff_pseudocode}

\begin{algorithm}[ht!]
\begin{lstlisting}
def module_diff(m1, m2):
  // m1 and m2 are DAG representations of the input
  // models. DAG nodes are torch.nn.module layers
  // (e.g., Linear, Conv2D). An edge between two
  // nodes indicates dataflow. 
  // We want to compute the diff, i.e., the nodes and
  // edges to remove and add to convert m1 to m2.
  
  // Compute hash tables of nodes/edges for m1 and m2
  // where values are node / edge lists sorted in
  // topological order. The hash of an edge is the
  // hash of its end points.
  N1, E1 = generate_hash_table(m1)
  N2, E2 = generate_hash_table(m2)
  
  // Iterate over E1: if a hash exists in E2,
  // greedily match each edge in two edge lists.
  // Before deciding on a matching, check the nodes
  // in these edges and only commit when 
  // corresponding nodes have the same matched
  // status. Matching a node in m1 with more than one
  // node in m2 is not allowed.
  Matches_N, Matches_E = {}, {}
  for hash in E1:
    es1 = E1[hash], es2 = E2[hash]
    for e1 in es1:
      for e2 in es2:
        if check(e1, e2):
          e1[0].matched, e1[1].matched = True, True
          e2[0].matched, e2[1].matched = True, True
          Matches_N.add((e1[0],e2[0]), (e1[1],e2[1]))
          Matches_E.add((e1,e2))
          E2[hash].drop(e2)
      es2 = E2[hash]  
      
  // Match nodes that do not belong to common edges.
  for hash in N1:
    ns1 = [n1 in N1[hash] if n1.matched = False]
    ns2 = [n2 in N2[hash] if n2.matched = False]
    for i in range(min(len(ns1, ns2))):
        ns1[i].matched, ns2[i].matched = True, True
        Matches_N.add((ns1[i], ns2[i]))
        
  // Sort Matches_N/E by topological order of nodes /
  // edges in m1 and remove inverse matches of node /
  // edges: if n1 in m1 is matched with an n2 in m2
  // after one of n1's preceding nodes in m1 has been
  // matched with a node that appears later than the
  // suggested n2.
  // E.g., A-B-A-C and A-B-C-A should have a node
  // matching of only {A, B, C (or A)}.
  Matches_N = filter(sort(Matches_N))
  Matches_E = filter(sort(Matches_E))

  // Add_N/E are the unmatched nodes/edges in m2.
  // Del_N/E are the unmatched nodes/edges in m1.
  Add_E = E2.difference(e2 in Matches_E)
  Del_E = E1.difference(e1 in Matches_E)
  Add_N = N2.difference(n2 in Matches_N)
  Del_N = N1.difference(n1 in Matches_N)
  return Add_E, Add_N, Del_E, Del_N
\end{lstlisting}
\caption{Peudocode for \texttt{diff} between two models $m1$ and $m2$.}
\label{alg:graph_diff}
\end{algorithm}

%% file: main.bbl
\begin{thebibliography}{46}
\providecommand{\natexlab}[1]{#1}
\providecommand{\url}[1]{\texttt{#1}}
\expandafter\ifx\csname urlstyle\endcsname\relax
  \providecommand{\doi}[1]{doi: #1}\else
  \providecommand{\doi}{doi: \begingroup \urlstyle{rm}\Url}\fi

\bibitem[Abadi et~al.(2016)Abadi, Barham, Chen, Chen, Davis, Dean, Devin,
  Ghemawat, Irving, Isard, et~al.]{abadi2016tensorflow}
Abadi, M., Barham, P., Chen, J., Chen, Z., Davis, A., Dean, J., Devin, M.,
  Ghemawat, S., Irving, G., Isard, M., et~al.
\newblock {TensorFlow: A System for Large-Scale Machine Learning}.
\newblock In \emph{12th USENIX Symposium on Operating Systems Design and
  Implementation}, pp.\  265--283, 2016.

\bibitem[Ben~Zaken et~al.(2022)Ben~Zaken, Goldberg, and
  Ravfogel]{ben-zaken-etal-2022-bitfit}
Ben~Zaken, E., Goldberg, Y., and Ravfogel, S.
\newblock {{B}it{F}it: Simple Parameter-efficient Fine-tuning for
  Transformer-based Masked Language-models}.
\newblock In \emph{Proceedings of the 60th Annual Meeting of the Association
  for Computational Linguistics (Volume 2: Short Papers)}, pp.\  1--9, Dublin,
  Ireland, May 2022. Association for Computational Linguistics.

\bibitem[Bommasani et~al.(2021)Bommasani, Hudson, Adeli, Altman, Arora, von
  Arx, Bernstein, Bohg, Bosselut, Brunskill,
  et~al.]{bommasani2021opportunities}
Bommasani, R., Hudson, D.~A., Adeli, E., Altman, R., Arora, S., von Arx, S.,
  Bernstein, M.~S., Bohg, J., Bosselut, A., Brunskill, E., et~al.
\newblock {On the Opportunities and Risks of Foundation Models}.
\newblock \emph{arXiv preprint arXiv:2108.07258}, 2021.

\bibitem[Bonawitz et~al.(2019)Bonawitz, Eichner, Grieskamp, Huba, Ingerman,
  Ivanov, Kiddon, Kone{\v{c}}n{\`y}, Mazzocchi, McMahan,
  et~al.]{bonawitz2019towards}
Bonawitz, K., Eichner, H., Grieskamp, W., Huba, D., Ingerman, A., Ivanov, V.,
  Kiddon, C., Kone{\v{c}}n{\`y}, J., Mazzocchi, S., McMahan, B., et~al.
\newblock {Towards Federated Learning at Scale: System Design}.
\newblock \emph{Proceedings of Machine Learning and Systems}, 1:\penalty0
  374--388, 2019.

\bibitem[Brown et~al.(2020)Brown, Mann, Ryder, Subbiah, Kaplan, Dhariwal,
  Neelakantan, Shyam, Sastry, Askell, et~al.]{brown2020language}
Brown, T., Mann, B., Ryder, N., Subbiah, M., Kaplan, J.~D., Dhariwal, P.,
  Neelakantan, A., Shyam, P., Sastry, G., Askell, A., et~al.
\newblock {Language Models are Few-Shot Learners}.
\newblock \emph{Advances in Neural Information Processing Systems},
  33:\penalty0 1877--1901, 2020.

\bibitem[Cao et~al.(2021)Cao, Aziz, and Titov]{model_editor}
Cao, N.~D., Aziz, W., and Titov, I.
\newblock Editing factual knowledge in language models.
\newblock In \emph{EMNLP (1)}, pp.\  6491--6506, 2021.
\newblock URL \url{https://doi.org/10.18653/v1/2021.emnlp-main.522}.

\bibitem[Deng et~al.(2009)Deng, Dong, Socher, Li, Li, and
  Fei-Fei]{deng2009imagenet}
Deng, J., Dong, W., Socher, R., Li, L.-J., Li, K., and Fei-Fei, L.
\newblock {ImageNet: A Large-Scale Hierarchical Image Database}.
\newblock In \emph{2009 IEEE conference on computer vision and pattern
  recognition}, pp.\  248--255. Ieee, 2009.

\bibitem[Devlin et~al.(2018)Devlin, Chang, Lee, and Toutanova]{devlin2018bert}
Devlin, J., Chang, M.-W., Lee, K., and Toutanova, K.
\newblock {BERT: Pre-Training of Deep Bidirectional Transformers for Language
  Understanding}.
\newblock \emph{arXiv preprint arXiv:1810.04805}, 2018.

\bibitem[Du et~al.(2022)Du, Huang, Dai, Tong, Lepikhin, Xu, Krikun, Zhou, Yu,
  Firat, et~al.]{du2022glam}
Du, N., Huang, Y., Dai, A.~M., Tong, S., Lepikhin, D., Xu, Y., Krikun, M.,
  Zhou, Y., Yu, A.~W., Firat, O., et~al.
\newblock {GLaM: Efficient Scaling of Language Models with Mixture-of-Experts}.
\newblock In \emph{International Conference on Machine Learning}, pp.\
  5547--5569. PMLR, 2022.

\bibitem[Fedus et~al.(2022)Fedus, Zoph, and Shazeer]{fedus2022switch}
Fedus, W., Zoph, B., and Shazeer, N.
\newblock {Switch Transformers: Scaling to Trillion Parameter Models with
  Simple and Efficient Sparsity}.
\newblock \emph{The Journal of Machine Learning Research}, 23\penalty0
  (1):\penalty0 5232--5270, 2022.

\bibitem[Guo et~al.(2021{\natexlab{a}})Guo, Rush, and
  Kim]{guo-etal-2021-parameter}
Guo, D., Rush, A., and Kim, Y.
\newblock {Parameter-Efficient Transfer Learning with Diff Pruning}.
\newblock In \emph{Proceedings of the 59th Annual Meeting of the Association
  for Computational Linguistics and the 11th International Joint Conference on
  Natural Language Processing (Volume 1: Long Papers)}, pp.\  4884--4896,
  Online, 2021{\natexlab{a}}. Association for Computational Linguistics.

\bibitem[Guo et~al.(2021{\natexlab{b}})Guo, Hu, and Hu]{guo2021mistify}
Guo, P., Hu, B., and Hu, W.
\newblock {Mistify: Automating DNN Model Porting for On-Device Inference at the
  Edge}.
\newblock In \emph{18th USENIX Symposium on Networked Systems Design and
  Implementation (NSDI 21)}, pp.\  705--719, 2021{\natexlab{b}}.

\bibitem[Hao et~al.(2022)Hao, Awatramani, Hu, Mao, Chen, Cidon, Cidon, and
  Yang]{hao2022tale}
Hao, W., Awatramani, A., Hu, J., Mao, C., Chen, P.-C., Cidon, E., Cidon, A.,
  and Yang, J.
\newblock {A Tale of Two Models: Constructing Evasive Attacks on Edge Models}.
\newblock \emph{Proceedings of Machine Learning and Systems}, 4:\penalty0
  414--429, 2022.

\bibitem[Hase et~al.(2021)Hase, Diab, Celikyilmaz, Li, Kozareva, Stoyanov,
  Bansal, and Iyer]{hase}
Hase, P., Diab, M., Celikyilmaz, A., Li, X., Kozareva, Z., Stoyanov, V.,
  Bansal, M., and Iyer, S.
\newblock Do language models have beliefs? methods for detecting, updating, and
  visualizing model beliefs, 2021.
\newblock URL \url{https://arxiv.org/abs/2111.13654}.

\bibitem[He et~al.(2016)He, Zhang, Ren, and Sun]{he2016deep}
He, K., Zhang, X., Ren, S., and Sun, J.
\newblock {Deep Residual Learning for Image Recognition}.
\newblock In \emph{Proceedings of the IEEE Conference on Computer Vision and
  Pattern Recognition}, pp.\  770--778, 2016.

\bibitem[Houlsby et~al.(2019)Houlsby, Giurgiu, Jastrzebski, Morrone,
  de~Laroussilhe, Gesmundo, Attariyan, and
  Gelly]{Houlsby2019ParameterEfficientTL}
Houlsby, N., Giurgiu, A., Jastrzebski, S., Morrone, B., de~Laroussilhe, Q.,
  Gesmundo, A., Attariyan, M., and Gelly, S.
\newblock {Parameter-Efficient Transfer Learning for NLP}.
\newblock In \emph{ICML}, 2019.

\bibitem[Howard et~al.(2017)Howard, Zhu, Chen, Kalenichenko, Wang, Weyand,
  Andreetto, and Adam]{howard2017mobilenets}
Howard, A.~G., Zhu, M., Chen, B., Kalenichenko, D., Wang, W., Weyand, T.,
  Andreetto, M., and Adam, H.
\newblock {MobileNets: Efficient Convolutional Neural Networks for Mobile
  Vision Applications}.
\newblock \emph{arXiv preprint arXiv:1704.04861}, 2017.

\bibitem[Hu et~al.(2022)Hu, yelong shen, Wallis, Allen-Zhu, Li, Wang, Wang, and
  Chen]{hu2022lora}
Hu, E.~J., yelong shen, Wallis, P., Allen-Zhu, Z., Li, Y., Wang, S., Wang, L.,
  and Chen, W.
\newblock {Lo{RA}: Low-Rank Adaptation of Large Language Models}.
\newblock In \emph{International Conference on Learning Representations}, 2022.

\bibitem[Hu et~al.(2020)Hu, Zou, Xia, Jin, Tao, Liu, Zhang, and
  Zhang]{hu2020delta}
Hu, Z., Zou, X., Xia, W., Jin, S., Tao, D., Liu, Y., Zhang, W., and Zhang, Z.
\newblock {Delta-DNN: Efficiently Compressing Deep Neural Networks via
  Exploiting Floats Similarity}.
\newblock In \emph{49th International Conference on Parallel Processing-ICPP},
  pp.\  1--12, 2020.

\bibitem[Huang et~al.(2017)Huang, Liu, Van Der~Maaten, and
  Weinberger]{huang2017densely}
Huang, G., Liu, Z., Van Der~Maaten, L., and Weinberger, K.~Q.
\newblock {Densely Connected Convolutional Networks}.
\newblock In \emph{Proceedings of the IEEE conference on computer vision and
  pattern recognition}, pp.\  4700--4708, 2017.

\bibitem[{HuggingFace}()]{huggingface_model_hub}
{HuggingFace}.
\newblock {HuggingFace Model Hub}.
\newblock \url{https://huggingface.co/models}.

\bibitem[Igor(1998)]{LZMA}
Igor, P.
\newblock {The Algorithm: Lempel-Ziv-Markov Chain}.
\newblock 1998.

\bibitem[Jouppi et~al.(2017)Jouppi, Young, Patil, Patterson, Agrawal, Bajwa,
  Bates, Bhatia, Boden, Borchers, et~al.]{jouppi2017datacenter}
Jouppi, N.~P., Young, C., Patil, N., Patterson, D., Agrawal, G., Bajwa, R.,
  Bates, S., Bhatia, S., Boden, N., Borchers, A., et~al.
\newblock {In-Datacenter Performance Analysis of a Tensor Processing Unit}.
\newblock In \emph{Proceedings of the 44th annual international symposium on
  computer architecture}, pp.\  1--12, 2017.

\bibitem[Jumper et~al.(2021)Jumper, Evans, Pritzel, Green, Figurnov,
  Ronneberger, Tunyasuvunakool, Bates, {\v{Z}}{\'\i}dek, Potapenko,
  et~al.]{jumper2021highly}
Jumper, J., Evans, R., Pritzel, A., Green, T., Figurnov, M., Ronneberger, O.,
  Tunyasuvunakool, K., Bates, R., {\v{Z}}{\'\i}dek, A., Potapenko, A., et~al.
\newblock {Highly Accurate Protein Structure Prediction with AlphaFold}.
\newblock \emph{Nature}, 596\penalty0 (7873):\penalty0 583--589, 2021.

\bibitem[Kang et~al.(2020)Kang, Raghavan, Bailis, and Zaharia]{kang2020model}
Kang, D., Raghavan, D., Bailis, P., and Zaharia, M.
\newblock {Model Assertions for Monitoring and Improving ML Models}.
\newblock \emph{Proceedings of Machine Learning and Systems}, 2:\penalty0
  481--496, 2020.

\bibitem[Kone{\v{c}}n{\`y} et~al.(2016)Kone{\v{c}}n{\`y}, McMahan, Yu,
  Richt{\'a}rik, Suresh, and Bacon]{konevcny2016federated}
Kone{\v{c}}n{\`y}, J., McMahan, H.~B., Yu, F.~X., Richt{\'a}rik, P., Suresh,
  A.~T., and Bacon, D.
\newblock {Federated Learning: Strategies for Improving Communication
  Efficiency}.
\newblock \emph{arXiv preprint arXiv:1610.05492}, 2016.

\bibitem[McMahan et~al.(2017)McMahan, Moore, Ramage, Hampson, and
  y~Arcas]{mcmahan2017communication}
McMahan, B., Moore, E., Ramage, D., Hampson, S., and y~Arcas, B.~A.
\newblock {Communication-Efficient Learning of Deep Networks from Decentralized
  Data}.
\newblock In \emph{Artificial intelligence and statistics}, pp.\  1273--1282.
  PMLR, 2017.

\bibitem[Miao et~al.(2016)Miao, Li, Davis, and Deshpande]{miao2016modelhub}
Miao, H., Li, A., Davis, L.~S., and Deshpande, A.
\newblock {ModelHub: Towards Unified Data and Lifecycle Management for Deep
  Learning}.
\newblock \emph{arXiv preprint arXiv:1611.06224}, 2016.

\bibitem[Mitchell et~al.(2021)Mitchell, Lin, Bosselut, Finn, and
  Manning]{mitchell2021fast}
Mitchell, E., Lin, C., Bosselut, A., Finn, C., and Manning, C.~D.
\newblock Fast model editing at scale.
\newblock \emph{CoRR}, 2021.
\newblock URL \url{https://arxiv.org/pdf/2110.11309.pdf}.

\bibitem[Moradi \& Samwald(2021)Moradi and
  Samwald]{moradi-samwald-2021-evaluating}
Moradi, M. and Samwald, M.
\newblock {Evaluating the Robustness of Neural Language Models to Input
  Perturbations}.
\newblock In \emph{Proceedings of the 2021 Conference on Empirical Methods in
  Natural Language Processing}, pp.\  1558--1570. Association for Computational
  Linguistics, 2021.

\bibitem[Murshed et~al.(2021)Murshed, Murphy, Hou, Khan, Ananthanarayanan, and
  Hussain]{murshed2021machine}
Murshed, M.~S., Murphy, C., Hou, D., Khan, N., Ananthanarayanan, G., and
  Hussain, F.
\newblock {Machine Learning at the Network Edge: A Survey}.
\newblock \emph{ACM Computing Surveys (CSUR)}, 54\penalty0 (8):\penalty0 1--37,
  2021.

\bibitem[Paszke et~al.(2019)Paszke, Gross, Massa, Lerer, Bradbury, Chanan,
  Killeen, Lin, Gimelshein, Antiga, et~al.]{paszke2019pytorch}
Paszke, A., Gross, S., Massa, F., Lerer, A., Bradbury, J., Chanan, G., Killeen,
  T., Lin, Z., Gimelshein, N., Antiga, L., et~al.
\newblock {PyTorch: An Imperative Style, High-Performance Deep Learning
  Library}.
\newblock \emph{Advances in Neural Information Processing Systems}, 32, 2019.

\bibitem[Polino et~al.(2018)Polino, Pascanu, and Alistarh]{polino2018model}
Polino, A., Pascanu, R., and Alistarh, D.
\newblock {Model Compression via Distillation and Quantization}.
\newblock \emph{arXiv preprint arXiv:1802.05668}, 2018.

\bibitem[Pratt(1992)]{pratt1992discriminability}
Pratt, L.~Y.
\newblock {Discriminability-based Transfer between Neural Networks}.
\newblock \emph{Advances in Neural Information Processing Systems}, 5, 1992.

\bibitem[Qiu et~al.(2022)Qiu, Vavelidou, Li, Pergament, Warden, Chinchali,
  Asgar, and Katti]{mlexray}
Qiu, H., Vavelidou, I., Li, J., Pergament, E., Warden, P., Chinchali, S.,
  Asgar, Z., and Katti, S.
\newblock {ML-EXray: Visibility into ML Deployment on the Edge}.
\newblock \emph{Proceedings of Machine Learning and Systems}, 4:\penalty0
  337--351, 2022.

\bibitem[Rebuffi et~al.(2017)Rebuffi, Bilen, and Vedaldi]{rebuffi2017learning}
Rebuffi, S.-A., Bilen, H., and Vedaldi, A.
\newblock {Learning Multiple Visual Domains with Residual Adapters}.
\newblock \emph{Advances in Neural Information Processing Systems}, 30, 2017.

\bibitem[Reed et~al.(2022)Reed, DeVito, He, Ussery, and Ansel]{reed2022torch}
Reed, J., DeVito, Z., He, H., Ussery, A., and Ansel, J.
\newblock {torch. fx: Practical Program Capture and Transformation for Deep
  Learning in Python}.
\newblock \emph{Proceedings of Machine Learning and Systems}, 4:\penalty0
  638--651, 2022.

\bibitem[Ribeiro \& Lundberg(2022)Ribeiro and Lundberg]{adatest}
Ribeiro, M.~T. and Lundberg, S.
\newblock {Adaptive Testing and Debugging of {NLP} Models}.
\newblock In \emph{Proceedings of the 60th Annual Meeting of the Association
  for Computational Linguistics (Volume 1: Long Papers)}, pp.\  3253--3267,
  Dublin, Ireland, 2022. Association for Computational Linguistics.

\bibitem[Ribeiro et~al.(2020)Ribeiro, Wu, Guestrin, and Singh]{checklist}
Ribeiro, M.~T., Wu, T., Guestrin, C., and Singh, S.
\newblock {Beyond Accuracy: Behavioral Testing of {NLP} Models with
  {C}heck{L}ist}.
\newblock In \emph{Proceedings of the 58th Annual Meeting of the Association
  for Computational Linguistics}, pp.\  4902--4912, Online, 2020. Association
  for Computational Linguistics.

\bibitem[Robinson \& Cherry(1967)Robinson and Cherry]{RLE}
Robinson, A. and Cherry, C.
\newblock {Results of a Prototype Television Bandwidth Compression Scheme}.
\newblock \emph{Proceedings of the IEEE}, 55\penalty0 (3):\penalty0 356--364,
  1967.
\newblock \doi{10.1109/PROC.1967.5493}.

\bibitem[Ruder(2017)]{ruder2017overview}
Ruder, S.
\newblock {An Overview of Multi-Task Learning in Deep Neural Networks}.
\newblock \emph{arXiv preprint arXiv:1706.05098}, 2017.

\bibitem[Wang et~al.(2018)Wang, Singh, Michael, Hill, Levy, and
  Bowman]{wang2018glue}
Wang, A., Singh, A., Michael, J., Hill, F., Levy, O., and Bowman, S.~R.
\newblock {GLUE: A Multi-Task Benchmark and Analysis Platform for Natural
  Language Understanding}.
\newblock \emph{arXiv preprint arXiv:1804.07461}, 2018.

\bibitem[Weiss et~al.(2016)Weiss, Khoshgoftaar, and Wang]{weiss2016survey}
Weiss, K., Khoshgoftaar, T.~M., and Wang, D.
\newblock {A Survey of Transfer Learning}.
\newblock \emph{Journal of Big data}, 3\penalty0 (1):\penalty0 1--40, 2016.

\bibitem[Wolf et~al.(2019)Wolf, Debut, Sanh, Chaumond, Delangue, Moi, Cistac,
  Rault, Louf, Funtowicz, et~al.]{wolf2019huggingface}
Wolf, T., Debut, L., Sanh, V., Chaumond, J., Delangue, C., Moi, A., Cistac, P.,
  Rault, T., Louf, R., Funtowicz, M., et~al.
\newblock {Huggingface's Transformers: State-of-the-Art Natural Language
  Processing}.
\newblock \emph{arXiv preprint arXiv:1910.03771}, 2019.

\bibitem[Zaken et~al.(2021)Zaken, Ravfogel, and Goldberg]{zaken2021bitfit}
Zaken, E.~B., Ravfogel, S., and Goldberg, Y.
\newblock {BitFit: Simple Parameter-Efficient Fine-Tuning for Transformer-Based
  Masked Language Models}.
\newblock \emph{arXiv preprint arXiv:2106.10199}, 2021.

\bibitem[Zhou et~al.(2022)Zhou, Wei, Zhang, and Sun]{pets}
Zhou, Z., Wei, X., Zhang, J., and Sun, G.
\newblock {{PetS}: A Unified Framework for {Parameter-Efficient} Transformers
  Serving}.
\newblock In \emph{2022 USENIX Annual Technical Conference (USENIX ATC 22)},
  pp.\  489--504, Carlsbad, CA, 2022. USENIX Association.

\end{thebibliography}
